\documentclass{article}

\usepackage{arxiv}

\usepackage[utf8]{inputenc} % allow utf-8 input
\usepackage[T1]{fontenc}    % use 8-bit T1 fonts
\usepackage{hyperref}       % hyperlinks
\usepackage{url}            % simple URL typesetting
\usepackage{booktabs}       % professional-quality tables
\usepackage{amsfonts}       % blackboard math symbols
\usepackage{nicefrac}       % compact symbols for 1/2, etc.
\usepackage{microtype}      % microtypography
\usepackage{lipsum}		% Can be removed after putting your text content
\usepackage{graphicx}
\usepackage{doi}
\usepackage{amsmath}
\usepackage{fontawesome}
\usepackage{hyperref}

\title{Orthogonal Transforms in Neural Networks\\ Amount to Effective Regularization}

\author{ 
    Krzysztof Zając \\
	Wrocław University of Science and Technology \\
	\texttt{krzysztof.zajac@pwr.edu.pl} \\
    \And
	Wojciech Sopot \\
	Wrocław University of Science and Technology \\
	\texttt{wojciech.sopot@pwr.edu.pl} \\
	\And
	Paweł Wachel \\
	Wrocław University of Science and Technology \\
	\texttt{pawel.wachel@pwr.edu.pl} \\
}

\renewcommand{\shorttitle}{Orthogonal Transforms in Neural Networks}

\hypersetup{
pdftitle={Orthogonal Transforms in Neural Networks Amount to Effective Regularization: Pre-Print},
pdfsubject={},
pdfauthor={Krzysztof Zając, Wojciech Sopot, Paweł Wachel},
pdfkeywords={Neural Networks, Nonlinear Dynamics, Orthogonal Transform, System Identification},
}

\begin{document}
\maketitle

\begin{abstract}
	We consider applications of neural networks in nonlinear system identification and formulate a hypothesis that adjusting general network structure by incorporating frequency information or other known orthogonal transform, should result in an efficient neural network retaining its universal properties. We show that such a structure is a universal approximator and that using any orthogonal transform in a proposed way implies regularization during training by adjusting the learning rate of each parameter individually. We empirically show in particular, that such a structure, using the Fourier transform, outperforms equivalent models without orthogonality support.
\end{abstract}


\section{Introduction}
Neural networks are a very general type of parametric model, capable of learning various relationships between variables in various modalities. This generality is one of their greatest strengths, but special cases designed for specific applications exist. Those include convolutional neural networks for processing data with spatial relationships and recurrent neural networks for sequential data. One of the causes of the good performance of specialized models in specific cases is that those networks have useful inductive \textit{a priori} knowledge about the task added during the architecture development processes. Those biases can be very general, such as translation invariance for convolutional networks or causality and memory introduced in recurrent neural networks, particularly long-short term memory network \cite{lstm} and gated-recurrent unit \cite{gru}. 

Among different tasks of modern system modelling and identification, one can consider the problem of estimating the system's output, conditioned on its past values and excitation. In this context,  two classes of problems are distinguished: \textit{simulation modelling} and \textit{predictive modelling} \cite{nonlinear_system_identification}. \textit{Simulation modelling} is a task in which outputs are predicted based only on the input signal, while \textit{prediction modelling} requires measurements of the system trajectory to predict future states (hybrid approaches are also possible). This paper considers simulation modelling, which can be framed as an optimization problem of $n$-step ahead prediction based on $m$ samples of the input sequence. The training dataset contains measurements of input and output signals from the system, which are grouped into windows of fixed length.

Dynamical system identification, as a sequence modelling task, has several similarities to natural language processing. However, the inductive biases are different for both problems. In language, positional information is much more relevant than in dynamical systems modelling. In turn, for the dynamical system, it is possible to derive useful biases from physical insight. We hypothesise that adding frequency information to the network's structure will be useful inductive knowledge for dynamical system identification in a simulation setting. A similar approach was used to derive Fourier neural operators, which achieve good performance in fluid dynamics using predictive modelling \cite{original_fno,adaptive_fno}, while still being very general and applicable to various kinds of physical or engineering systems. Different orthogonal transforms, such as wavelet transforms are also considered in literature \cite{wavelet_no,wno_physics}.
The number of methods developed for the identification of nonlinear systems is very large. Some of them include domain knowledge \cite{nonlinear_physical_models,QBLA,black_and_grey_box}, while others are very general and require only very mild assumptions about the nature of modelled systems \cite{volterra_series,reservoir_computation,pde_modelling,input_injection,inf_memory_nonlinear_systems}. One of the most successful classes of models applied to the identification of nonlinear dynamics are neural networks \cite{ribeiro_deep_rnn,cnn_system_id,transformer_physics}. Nevertheless, even given the generality of the network structure, many specialized networks have been developed specifically for nonlinear dynamics and usually achieve better results than the less specialized structures \cite{dyno_net,state_space_encoder}. Often those specialized networks are built using general \textit{a priori} knowledge about the nature of the system; sometimes, they have knowledge about physical equations baked into the architecture \cite{physics_informed_networks}.
Our hypothesis leads to the formulation of a network structure resembling that of Fourier Neural Operator \cite{original_fno}. Such a model processes the system input signal in parallel in the time and frequency domain. We theoretically analyse this structure, showing that both branches are universal approximators, so such a structure also retains this property originally proved for a feed-forward network. We also show how adding such transforms impacts the learning process, namely that it effectively scales the gradient, for each parameter separately. Both those results hold for any orthogonal transform. We implement such a dual model and empirically apply it for simulation modelling, where it outperforms models without added frequency information\footnote{Model implementation and datasets used for experiments are open-source, available at \\ \hspace{2pt} \href{https://github.com/cyber-physical-systems-group/orthogonal-neural-networks}{\faGithubSquare \hspace{0.5pt} \texttt{https://github.com/cyber-physical-systems-group/orthogonal-neural-networks}}}.
\section{Dual-Orthogonal Neural Network}
The investigated structure of a dual neural network with orthogonal transform is an extension of standard feed-forward network. It is designed to incorporate useful information about the input signals by transforming it into a different basis. The network is a layered structure consisting of blocks, which can be arranged in any way, as long as the correct length of the input and output is preserved. The dual block is a building block of such a network and can be arranged into a deep network or combined with other blocks. The input of the block is a sequence, which can be a series of measurements of a dynamical system or some other time-dependent variable. Its output is also a sequence, though its length does not need to be preserved. In general, the input can be multi-dimensional (as well as the output) and transformations between dimensions are possible using this type of network. In experiments, we are using the Fourier transform, following the assumption that frequencies of the input signal will be useful for identification.
\subsection{Dual-Orthogonal Block}
Each dual block we consider consists of two parallel branches. One operates in the time domain and is a linear block processing the input sequence, whereas the other one is designed to focus on the space in a different basis, depending on the orthogonal transform used. Block combining more than two representations are also conceivable, but we leave them for future research. Outputs of both branches are added together and passed through a non-linear activation function $\sigma$. Any of the commonly used functions could be used, including \textit{Sigmoid}, \textit{ReLU} or \textit{Hyperbolic Tangent} \cite{activations}. Each block has four hyper-parameters: the length of the \textit{input} sequence, $S_i$, the length of the \textit{output} sequence it produces, $S_o$, and the input and output dimensionalities, $D_i$ and $D_o$, respectively. This allows adjusting the model to different types of dynamical systems, even those with a high number of state dimensions. The output of each time-branch can be expressed by equation 
\begin{equation}\label{time_branch_eq}
    \hat{h}_{l} = xW_l^{\mathsf{T}} + b_l,
\end{equation}
where $x \in \mathbb{R}^{S_i}$ is a row vector with length $S_i$, and $W_l$ is a matrix of learnable parameters with shape $[S_o \times S_i]$, and $b_l$ is a bias vector with length $S_o$. The output $\hat{h}_l \in \mathbb{R}^{S_o}$ is a row vector with length $S_o$. Elements of the time branch are denoted with subscript $l$ to distinguish them from the elements of the transformed branch, denoted with subscript $t$. The orthogonal block uses an orthogonal transform to convert the signal, applies learned linear transformation in this space and then converts back to the original domain
\begin{equation}\label{frequency_block}
    \hat{h}_t = \mathcal{T}^{-1} \left ( \mathcal{T}(x)W_t^{\mathsf{T}} + b_t \right ),
\end{equation}
where $\mathcal{T}$ denotes the orthogonal transform applied to the signal, and $\mathcal{T}^{-1}$ is its inverse. The matrix of parameters $W_t$ can be real or complex-valued. In general, the shape of matrices $W_l$ and $W_t$ is the same, but in some cases, such as Fourier transform certain parameters might be simplified. The output of each frequency-supported block is computed as the sum of representations produced by both branches with a nonlinear activation function $\sigma$, \textit{i.e.},
\begin{equation}\label{fsnn_output_eq}
    \hat{y} = \sigma(\hat{h}_l + \hat{h}_t).
\end{equation}
The structure of the dual-orthogonal neural network can be extended to the system with multiple-input multiple-output (MIMO) systems by extending the parameter matrices and using reshape operation. The input in the MIMO case is a matrix with shape $[S_i \times D_i]$, which is converted into a row vector of length $S_i D_i$, and the produced output is another matrix with shape $[S_o \times D_o]$. The matrices $W_l$ and $W_t$ are extended to shape $[S_oD_o \times S_iD_i]$. Bias vectors are also extended to contain the required parameters and have length $S_o D_o $. The output of the block in such a case is a vector with length $S_o D_o$, which can be rearranged into a matrix with the desired shape. The rearranging can be done before or after the aggregation and activation function because they are both element-wise operations.
\begin{figure}[h]
    \centerline{\includegraphics[width=\textwidth]{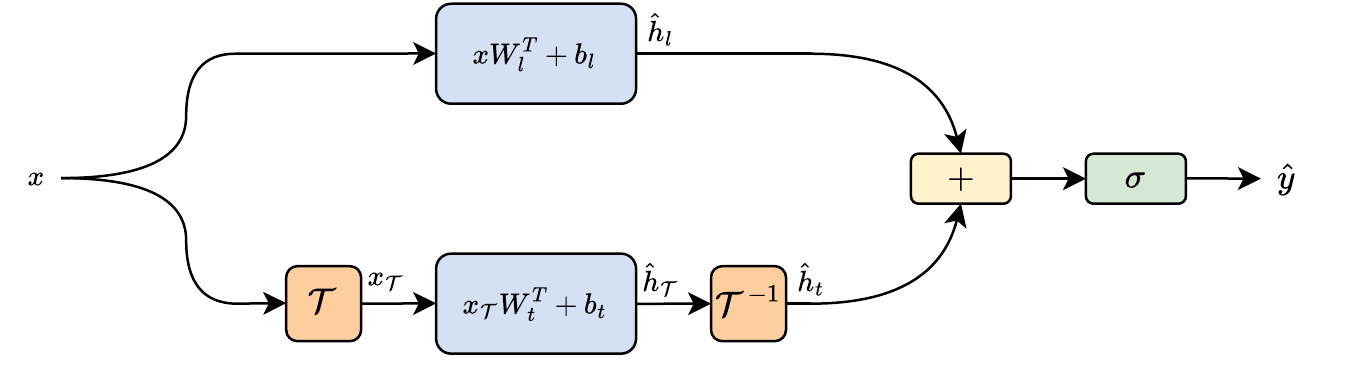}}
    \caption{Schematic representation of a dual-orthogonal block structure} \label{fsnn_schema}
\end{figure}
\subsection{Frequency-Supported Block}
In this study we are particularly focused on the dual block using Fourier transform, which will be further called \textit{frequency-supported block} and multiple such blocks, arranged sequentially will be called \textit{frequency-supported neural network}. The usage of the Fourier transform is dictated by the fact, that many practical dynamical systems can be modelled in frequency space effectively, so this transformation should be useful inductive bias. 
We use real-valued Fourier transform \cite{rfft_ieee}, so only the positive side of the spectrum is processed due to the assumption that such a model will process only real-valued signals. Due to the FFT algorithm used in implementation, it is most efficient on sequence lengths being a power of 2, \textit{cf.} \cite{fft_ieee}. In such case, the learned parameters are complex-valued in this branch. Using RFFT as $\mathcal{T}$ in formulation given by \eqref{frequency_block} allows us to reduce the number of parameters, due to our assumption about the real-valued signal. The matrix of parameters $W_t$ is complex-valued with shape $[1/2 \, S_o \times 1/2 \, S_i]$. It is smaller than in the time branch since the length of vector $x$ after applying the real-valued Fourier transform is half of its original length. Similarly, $b_t$ is a complex-valued vector of parameters with length $1/2 \, S_o$. As mentioned above, such a branch can also be extended to MIMO cases.
\subsection{$N$-Step Ahead Prediction}
The requirement for using the frequency-supported network is the availability of a training dataset, which needs to be composed of input and output measurements of the system of interest aligned in time with constant sampling between measurements. Moreover, the model structure requires the input and output to be of the known and constant length, which, in general, can be different. To simulate a dynamical system using our approach, the input signal needs to be split into windows, for which outputs will be predicted, where each window of inputs corresponds to a single window of outputs. In practice, short output windows and long input windows are most efficient, which is intuitively clear since longer input gives the model more information and shorter output makes error accumulation smaller. It is also possible to have different lengths in a sequential frequency-supported model, where the only requirement is that the output length of a given layer must match the input of the following layer. In the experiments, time windows were created using two parameters, $m$ for input length and $n$ for output length and time index $s$ (we denote index as sequence step $s$, while $t$ is used for transform), which was moved along the sequence of sampling times $S$. During training, the overlap is possible and sometimes useful so the same measurement can appear in different parts of the input or target sequence
\begin{equation} \label{training_set_builder}
    \{ (U_{(s-m) : s}, Y_{(s-n) : s}) \: | \: s \in S \}.
\end{equation}
Creating a training dataset as described in \eqref{training_set_builder} allows formulating a training procedure as a regression task, computing the mean-squared-error between model predictions and measured targets and optimizing the parameters using stochastic gradient descent or one of its variants \cite{sgd_overview}. $U$ and $Y$ in eq. \eqref{training_set_builder} denote measurements of the input signal and system output, respectively, which are indexed with the measurement time $t$, where the measurements of input signal $u(t)$ as the input to the first layer. For multi-dimensional systems, multiple vectors for excitation or output measurements can be included in the training dataset, all aligned using the same time index. 
\section{Theoretical Properties}
Feed-forward neural networks are known to have universal approximation property, which was originally proven in \cite{cybenko_proof} and extended in particular in \cite{hornik_universal_approximation,hornik_uaf_with_derivatives,uaf_non_sigmoid}. Informally, the universal approximation property guarantees that for any $n$-dimensional function $f$ from a given space, there exists a feed-forward neural network, $G(x)$, of the form given in \eqref{cybenko_torch_like}, such that $\lvert G(x) - f(x) \rvert < \epsilon$ for arbitrarily small $\epsilon>0$. In the case of simulation modelling of dynamical systems, the input to the network, $x \in \mathbb{R}^N$, is an input signal with finite time steps. Nevertheless, to guarantee this property, \textit{cf.} \cite{cybenko_proof}, the network is required to have an infinite number\footnote{For certain sophisticated activations, the number of units does not need to be infinite \cite{finite_size_uaf}. These models are however of lower practical potential.} of neurons (also called units) in the hidden layer with sigmoidal activation. Furthermore, it is required that
\begin{equation}\label{sigmoid_cybenko}
    \sigma(x) =
    \begin{cases}
    1 \quad \text{when} \quad x \rightarrow +\infty \\
    0 \quad \text{when} \quad x \rightarrow -\infty.\\
    \end{cases}
\end{equation}
The representation presented by \cite{cybenko_proof} allows showing that both time and orthogonal branches of dual-orthogonal network have universal approximation properties. For the time branch (\textit{i.e.}~orthogonal network with $\hat{h}_t\equiv 0$), this is straightforward, as the general form given in equation \eqref{cybenko_torch_like} expresses the same computation as the time branch
\begin{equation} \label{cybenko_torch_like}
    G(x) = \sigma (x W_l^{\mathsf{T}} + b_l) W_s + b_s.
\end{equation}
The learned parameters are in the hidden layer of the network, while $W_s$ and $b_s$ are additional readout parameters, which are also present in the original formulation \cite{cybenko_proof}.
\subsection{Orthogonal Branch as Universal Approximator}
The orthogonal branch (\textit{i.e.}~dual-orthogonal network with $\hat{h}_l\equiv 0$), is a universal approximator, which can be shown based on Cybenko's proof since it is possible to write it in a way equivalent to $G(x)$ network structure, utilizing the properties of the transformation matrix, given above. The general form of such branch in matrix notation is given by equation
\begin{equation} \label{frequency_branch_torch_like}
    G_{T}(x) = \sigma \left ((x T W_t^{\mathsf{T}} + b_t) T^{-1} \right ) W_s^{\mathsf{T}} + b_s,
\end{equation} where $T=[\phi_0, \phi_1, \phi_2, \cdots]$ denotes the transition matrix of orthogonal transform, with $\phi_i$ being $i$-th column vector of $\mathcal{T}$'s base. Such a matrix is infinitely wide, similarly to the hidden layer of a universal approximation network. When $\mathcal{T}$ is Fourier transform, the matrix $T$ is the DFT matrix \cite{dft_paper}. $T^{-1}$ corresponds to the inverse orthogonal transform, it is guaranteed to exist since $T$ is unitary. Given that parameters can take any value, it is possible to find matrix $W_{t}$ and complementing bias vector $b_t$, such that general form \eqref{frequency_branch_torch_like} is equivalent to universal feed-forward network \eqref{cybenko_torch_like}
\begin{eqnarray} \label{wf_and_bf}
    W_{f} & = & T W_{t}^{\mathsf{T}} T^{-1} \\
    b_f & = & b_t T^{-1}.
\end{eqnarray}
Those values can be computed analytically and they are presented in equations \eqref{wf_and_bf}. This property only holds for square matrices, since $T$ is always square by definition. After plugging in those values, the frequency branch has all the properties of a feed-forward network. 
\subsection{Gradients in Orthogonal Branch}
To show the influence of orthogonal transform on the learning process, we analyze the orthogonal branch at $n$-th layer (we set $W_l=0,b_l=0$), which is given by
\begin{equation}\label{orthogonal_branch}
    B^{(n)}(x)=\sigma\left({ \left(x T W_t^{\mathsf{T}}+ b_{t}\right) T^{-1}} \right).
\end{equation}
We calculate the derivative of the output $B^{(n)}(x)$ with respect to one of the matrix's $W_t$ parameters, which we denote $W_{t_{ab}}$,
\begin{equation} \label{frequency_branch_WS_like_deriv}
    \frac{\partial B^{(n)}(x)}{\partial W_{t_{ab}}} = x H_{ab}^{\mathsf{T}} \circ \sigma' \left({\left(x T W_t^{\mathsf{T}} + b_{t}\right) T^{-1} } \right),
\end{equation}
where elements of matrix $H_{ab}$ are defined as $H_{ab_{ij}}=\phi_{ia} \phi_{bj}$. In special case of Fourier basis ($T \equiv F^\mathsf{T}$, where $F$ is DFT matrix) $H_{ab_{ij}}=\frac{\omega^{ia+bj}}{N}$ and $\omega=e^{\frac{-2 \pi i}{N}}$. For any orthogonal transform, elements of $H_{ab}$ can be derived by writing the transformation as matrix multiplication. Derivative for bias vector will have an analogous form. For the learning process we assume that one of the variations of the stochastic gradient algorithm is used \cite{sgd_overview} to compute updates of parameters $\theta$ (weights and biases of the network) in batches. Updates are computed according to the general rule
\begin{equation}
    \theta_{e+1} = \theta_{e} - \alpha g_e,
\end{equation}
where $g_e$ denotes the gradient of the parameters with respect to the loss function at a certain step $e$ and $\alpha$ is a constant learning rate, typically set upfront. Values of $H_{ab}$ are determined per-parameter for every batch of training data, therefore such an approach can also be interpreted as adaptive per-parameter learning rate changing. 
In practice adaptive algorithms \cite{kingma2017adam} or learning rate scheduling is used \cite{automated_lr_schedule}. Learning rate schedule can be interpreted as using learning rate, which is a function of step $e$ or depends on past values of the loss $L$. In both cases, the learning rate is the same for all parameters, so our interpretation of the impact of orthogonality on the learning process is the same as for the constant learning rate.  For adaptive algorithms, such as Adam, the learning rate is set per parameter based on the history of its gradients
\begin{equation}
    \theta_{e+1} = \theta_{e} - \frac{\alpha}{\sqrt{\hat{\upsilon}_{e}} + \phi} \hat{m}_{e},
\end{equation}
where $\hat{m}_{e}$ and $\hat{\upsilon}_{e}$ are computed using decaying averages of past gradients, with decay speed controlled by additional hyper-parameters, which are similar to learning rate: $\beta_1$ and $\beta_2$. Values $\hat{m}_{e}$ and $\hat{\upsilon}_{e}$ are first and second moment respectively, \textit{i.e.},
\begin{equation}
    \hat{m}_{e+1}=\beta_1 \hat{m}_e + (1 - \beta_1)g_e,
\end{equation}
\begin{equation}
    \hat{\upsilon}_{e+1}=\beta_2 \hat{\upsilon}_{e} + (1 - \beta_2)g_e^2.
\end{equation}
Therefore, when each derivative is scaled by the influence of orthogonal transform, this can be interpreted as adaptive scaling of $\beta_1$ and $\beta_2$ learning rates, which are famously difficult to set. Every first-order optimization algorithm used in the neural network should behave similarly with respect to the orthogonal transform.

\section{Numerical Experiments}
Three numerical experiments were run to verify the hypothesis that adding frequency information to neural networks will be empirically beneficial. One consists of a toy problem with a static system, while the other two were benchmarks selected from system identification literature. 
Three core models were developed, for which a large grid search over the parameters was conducted. The frequency-supported neural network, (i), abbreviated FSNN, described in the previous section and consisting of a number of frequency-supported blocks stacked together. The FNN model, (ii) (\textit{i.e.} frequency neural network) consisting of only frequency blocks, which is a subset of the FSNN architecture, where $\hat{h}_l\equiv 0$. The final model (iii) was a regular feed-forward network processing the signal using delayed input measurements (we refer to it as MLP), which also is a subset of FSNN with $\hat{h}_t\equiv 0$. Additionally, selected state-of-art models were re-implemented and run on the same benchmark problems, or when available, the results were transferred from original papers. 
\subsection{Hyperparameter Search}
For all benchmarks and the three core architectures (FSNN, FNN and MLP) random search over a defined set of hyper-parameters was run, and later full grid search was run over the subset of hyperparameters, which were important for the model. The searched parameters were the following: number of input samples and number of predicted samples, number of hidden layers, and number of units in all layers. Some hyper-parameters were frozen and used for all models, such as the optimization algorithm \textit{Adam,} \cite{kingma2017adam}, and GeLU activation function, \textit{cf.} \cite{gelu_paper}.
The most important parameter is the number of input samples, especially for the models utilizing frequency information, since for shorter windows the amount of information about signal frequencies is lower, which makes it less useful. A smaller number of output samples effectively means the model needs to predict fewer time steps. A one-step-ahead prediction makes it usually more accurate, so the optimal value for all benchmarks was equal to one. However, using longer output windows could also be used effectively, especially when fast predictions are required by the application.
For the DynoNet model \cite{dyno_net}, reported values are a reproduction of the model using original code on different benchmarks, with the exception of the Wiener-Hammerstein system where the value from the original paper is reported. For the State Space Encoder reported results are also values from the original work \cite{state_space_encoder}, since reproducing the model was not possible due to very long training runtimes.
\subsection{Evaluation}
All the above algorithms were evaluated using root mean squared error (RMSE), which is a standard method of evaluation in regression problems. Physical units are added where possible. RMSE was computed as
\begin{equation} \label{rmse}
    RMSE(y,\hat{y}) = \sqrt{MSE(y, \hat{y})} = \sqrt{\frac{1}{N} \sum_{i=1}^{N}(y - \hat{y})^2}.
\end{equation}
Additionally, a normalized mean squared error was also evaluated, \textit{i.e.}~the ratio of RMSE and standard deviation of the predicted value (reported as a percentage)
\begin{equation}
    NRMSE(y, \hat{y}) = \frac{RMSE(y, \hat{y})}{\sigma_y}.
\end{equation}
\subsection{Static System with Frequency Input}
A simple static affine system with an input signal consisting of pure-tone sine waves was created to test FSNN structure, under conditions most suitable for it. The input signal was generated using a sum of frequencies drawn from uniform distribution: $\alpha \sim \mathcal{U}(-5, 5)$, which is given by eq. \eqref{forcing_toy_eq}  
\begin{equation} \label{forcing_toy_eq}
    u(t) = \sum_{i=1}^{5} sin(\alpha_i t).
\end{equation}
The output of the system was generated using two additional parameters, randomly drawn from the uniform distribution $\mathcal{U}(-5, 5)$, similar to the excitation frequencies. Those are denoted as $\beta_1,\beta_2$ in 
\label{toy_system_eq}
\begin{equation}
    f(u) = \beta_1 u(t) + \pi \beta_2.
\end{equation}
Results for this benchmark show the advantage of a two-branch structure over single-branch MLP and FNN models, where FSNN is able to achieve much better results, which are summarized in table \ref{table_ts}. During the experiments, the best-performing models were those with a low number of parameters, but larger models were also capable of achieving satisfactory results. 
DynoNet model achieving the best results on this benchmark had only one static layer, without the learnable dynamical operator, which also effectively made it a feed-forward network. However, the architecture is different than MLP, and the model could be easily obtained by performing a hyper-parameters search on the DynoNet model. Moreover, the DynoNet model is constructed in a way that allows one to easily infer the structure of the architecture given knowledge about the system since it can be decomposed into linear dynamical blocks and static nonlinearities \cite{dyno_net}. This allows for selecting good candidate architecture, however, such knowledge is not guaranteed in real-world situations.
\begin{table}[h!]
{\caption{Evaluation results for selected models on test dataset for a static affine system with frequency input, where $\#P$ is the number of parameters}\label{table_ts}}
    \centering
    \begin{tabular}{cccc}
    \hline
    \rule{0pt}{6pt}
        Model   & $\#P$& RMSE                 & NRMSE \\
    \hline
        FNN     & 1610 & 10.3 $\cdot 10^{-3}$ & 0.30\% \\ 
        MLP     & 2157 & 5.4  $\cdot 10^{-3}$ & 0.16\% \\
        FSNN    & 887  & 1.4  $\cdot 10^{-3}$ & 0.04\% \\
        DynoNet & 49   & 0.3  $\cdot 10^{-3}$ & 0.02\% \\
    \hline
    \end{tabular}
\end{table}
\begin{figure}[h!]
    \centerline{\includegraphics[height=2.5in, width=5in]{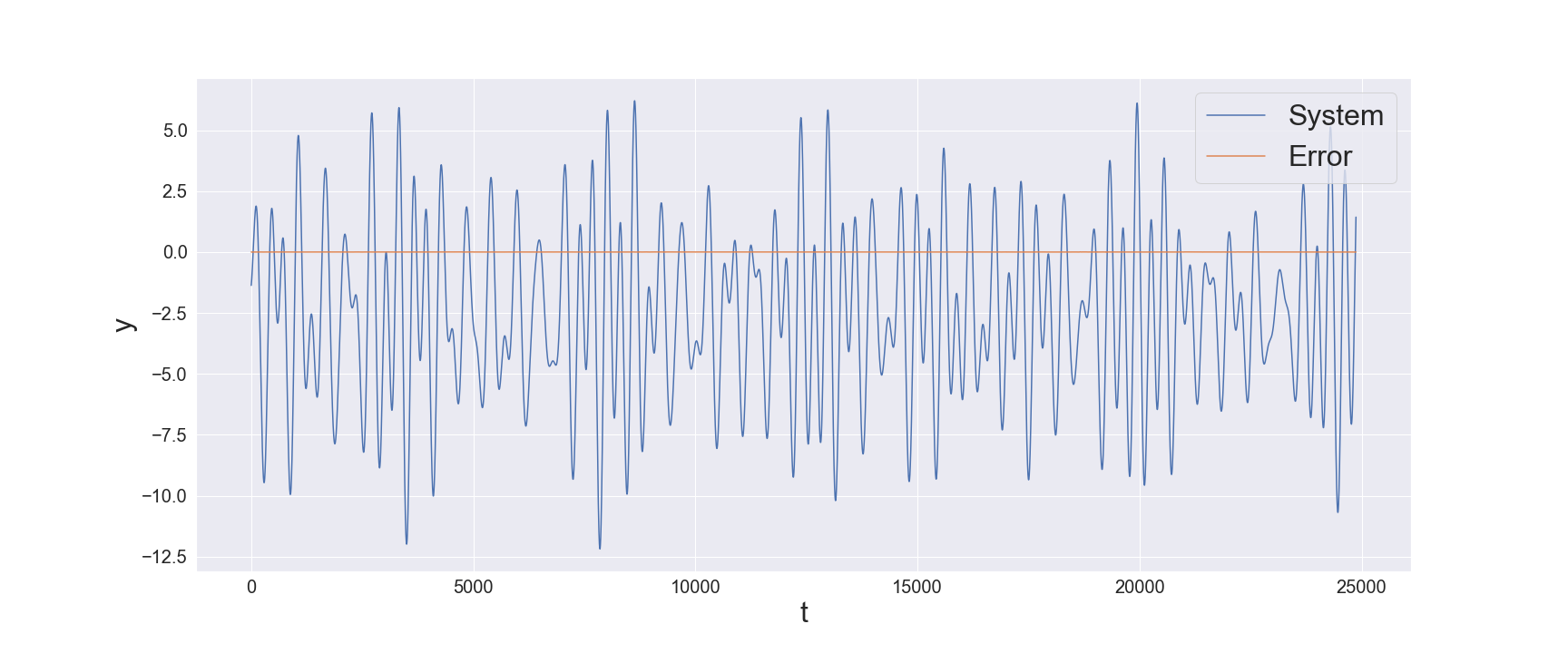}}
    \caption{Simulation error computed for best FSNN model on the test dataset for a static affine system with frequency input} \label{ts_predictions}
\end{figure}
\subsection{Wiener-Hammerstein Benchmark}
\begin{figure}[h!]
    \centerline{\includegraphics[height=2.5in, width=5in]{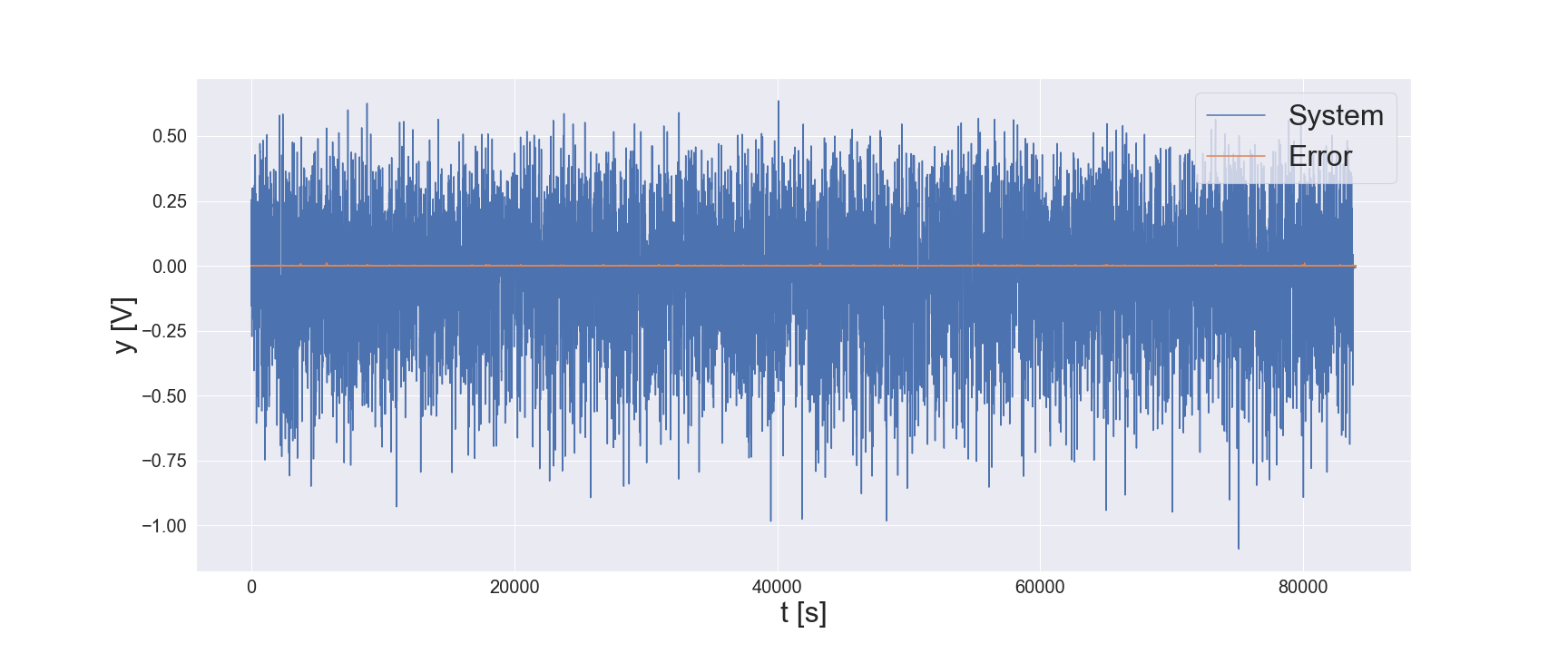}}
    \caption{Simulation error computed for best FSNN model on the test dataset for Wiener-Hammerstein benchmark} \label{wh_predictions}
\end{figure}

The Wiener-Hammerstein benchmark \cite{wiener_hammerstein_benchmark} is a well-known benchmark for the identification of nonlinear dynamics. It consists of two linear blocks and a static non-linearity, which were implemented using an electronic RLC circuit with a diode. The measurements of this system are used to create training and test datasets for the model.
\begin{table}[h!]
{\caption{Evaluation results for selected models on test dataset for Wiener-Hammerstein benchmark compared to selected results reported in literature}\label{table_wh}}
    \centering
    \begin{tabular}{cccc}
    \hline
    \rule{0pt}{6pt}
        Model   & $\#P$& RMSE                 & NRMSE \\
    \hline
        FNN                 & 7856    & 1.9 mV & 0.78\% \\
        DynoNet             & 63      & 1.2 mV & 0.50\% \\
        MLP                 & 1193    & 1.1 mV & 0.46\% \\
        Large MLP           & 1379841 & 0.9 mV & 0.38\% \\
        FSNN                & 1591    & 0.5 mV & 0.22\% \\
        State-Space Encoder & 21410   & 0.2 mV & 0.10\% \\
    \hline
    \end{tabular}
\end{table}
The results achieved by FSNN are not state-of-art, however, the structure performs significantly better than a plain MLP network on real-world data. For FSNN this performance was improved with longer input sequences, which allowed the model to access more frequency information. Evaluation results are reported in table \ref{table_wh} and for the selected model in figure \ref{wh_predictions}. For the MLP model, good results can be achieved with a wide range of hyper-parameters. The two reported results are on the two extremes of model size with a number of parameters different by three orders of magnitude, but have very similar performances. 
Both FSNN and MLP models are capable of achieving lower simulation error than the DynoNet model while having substantially fewer assumptions about the nature of modelled data. Additionally, it is worth noting that all listed models have normalized simulation error lower than $1\%$, which would be sufficient for most practical situations.
\subsection{Silverbox Benchmark}
Silverbox benchmark is an electronic implementation of the Duffing oscillator, which can be modelled using a second-order LTI system with a polynomial nonlinearity in the feedback \cite{silverbox_benchmark}. This type of system is challenging due to its nonlinearity. Experiments performed using this benchmark were conducted in the same way as all the others.
\begin{table}[h!]
{\caption{Evaluation results for selected models on test dataset for Silverbox benchmark compared to selected results reported in literature}\label{silverbox_table}}
    \centering
    \begin{tabular}{cccc}
    \hline
    \rule{0pt}{6pt}
        Model   & $\#P$& RMSE                 & NRMSE \\
    \hline
        FNN                 & 14192   & 4.1 mV & 7.69\% \\
        MLP                 & 37313   & 3.9 mV & 7.32\% \\
        DynoNet             & 81      & 2.9 mV & 5.39\% \\
        FSNN                & 69719   & 2.3 mV & 4.31\% \\
        State-Space Encoder & 19930   & 1.4 mV & 2.60\% \\
    \hline
    \end{tabular}
\end{table}
\par
\begin{figure}[hb!]
    \centerline{\includegraphics[height=2.5in, width=5in]{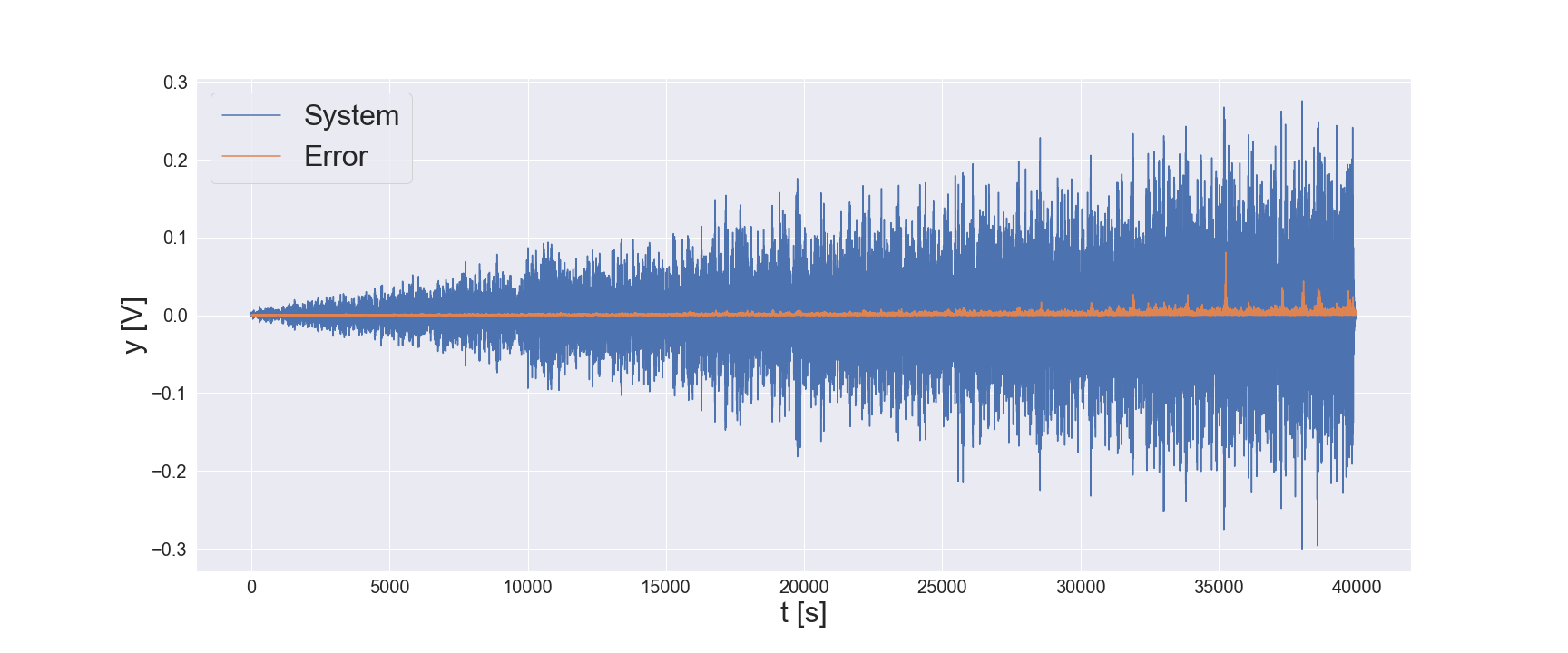}}
    \caption{Simulation error computed for best FSNN model on the test dataset for Silverbox benchmark} \label{sb_plot}
\end{figure}

 The structure of models applied to this benchmark cannot reflect the polynomial nonlinearity, which causes larger simulation errors when compared to the Wiener-Hammerstein benchmark. Moreover, models performing well on this benchmark tend to be larger then in previous cases, which also could be attributed to this nonlinearity. Results are reported in table \ref{silverbox_table} and for the selected model in figure \ref{sb_plot}.

\section{Discussion}
In conclusion, in the paper, we have shown that adding orthogonal transforms to neural networks does not restrict the generality of the network. Moreover we have demonstrated, that such modification can be interpreted as a method for regularization and initialization of the network. Experimental results, through the application of the Fourier transform, support the hypothesis that effective bases for system identification exist. Generally, it is likely to derive more useful bases for identification from physical insight or a priori knowledge about system structure. Therefore, the dual-orthogonal structure is a way of making neural models specialized for specific identification tasks and is worth further investigation.

\bibliographystyle{ieeetr}
\bibliography{references}

\end{document}